\documentclass[10pt,twocolumn,letterpaper]{article}

\usepackage[pagenumbers]{wacv} 

\definecolor{wacvblue}{rgb}{0.21,0.49,0.74}
\usepackage[pagebackref,breaklinks,colorlinks,allcolors=wacvblue]{hyperref}

\def\wacvPaperID{2620} 
\def\confName{WACV}
\def\confYear{2026}

\title{DiffRegCD: Integrated Registration and Change Detection with Diffusion Features
}

\author{Seyedehanita Madani \quad Rama Chellappa \quad Vishal M.
Patel \\
Johns Hopkins University\\
{\tt\small smadani4@jhu.edu} \quad {\tt\small rchella4@jhu.edu} \quad {\tt\small vpatel36@jhu.edu }
}

\begin{document}
\maketitle
\begin{abstract}
Change detection (CD) is critical in computer vision and remote sensing, with applications in monitoring, disaster response, and urban analysis. Most CD models assume co-registered inputs, but real imagery often suffers from parallax, viewpoint shifts, or long temporal gaps, leading to severe misalignment. Conventional register-then-detect pipelines and recent joint frameworks (e.g., BiFA, ChangeRD) remain limited: they rely on regression-only flow, global homographies, or synthetic perturbations that fail under large displacements.  We propose DiffRegCD, an integrated framework that couples dense registration and change detection. DiffRegCD reformulates correspondence as a Gaussian-smoothed classification task, delivering sub-pixel accuracy and stable training. It builds on frozen multi-scale features from a pretrained denoising diffusion model, which provide invariance to viewpoint and illumination variation. Supervision is enabled by controlled affine perturbations applied to standard CD datasets, yielding paired ground truth for both flow and change detection without pseudo-labels. Experiments on aerial (LEVIR-CD, DSIFN-CD, WHU-CD, SYSU-CD) and ground-level (VL-CMU-CD) datasets show that DiffRegCD outperforms recent baselines and remains robust under wide temporal and viewpoint variation, establishing diffusion features and classification-based correspondence as a strong foundation for integrated CD. The code is available at \href{https://github.com/Anita-Madani/DiffRegCD-Integrated-Registration-and-Change-Detection-with-Diffusion-Features}{GitHub}.

\end{abstract}

\vspace{-5mm}

\section{Introduction}
\label{sec:intro}

Change detection (CD) aims to identify semantic differences between two observations of the same scene across time. It is a central problem in computer vision and remote sensing, enabling applications in disaster assessment, environmental monitoring, urban growth analysis, and autonomous navigation. Despite progress, robust CD remains challenging because real-world inputs are rarely aligned. Variations in viewpoint, trajectory, illumination, and seasonal or multi-year gaps often introduce substantial misalignment, leading to spurious detections while obscuring true changes.  

A common strategy is to separate the problem into two stages: first register the images, then detect differences. Recent correspondence methods such as RAFT~\cite{teed2020raftrecurrentallpairsfield}, GMFlow~\cite{xu2022gmflowlearningopticalflow}, FlowFormer++~\cite{huang2022flowformertransformerarchitectureoptical}, LoFTR~\cite{sun2021loftrdetectorfreelocalfeature}, and RoMa~\cite{edstedt2023romarobustdensefeature} perform strongly in short-term matching but degrade under long baselines, where appearance shifts and structural modifications violate photometric consistency.  

In parallel, specialized CD models such as ChangeFormer~\cite{bandara2022transformerbasedsiamesenetworkchange} and DDPM-CD~\cite{bandara2022ddpm} achieve high accuracy on curated datasets (LEVIR-CD~\cite{Chen2020}, WHU-CD~\cite{ji2018fully}, DSIFN-CD \cite{ZHANG2020183} , SYSU-CD~\cite{shi21deeply}). However, they assume pre-aligned inputs and degrade sharply under misalignment caused by parallax, sensor variation, or temporal gaps.  

Several works have attempted to bridge registration and detection in a single model. BiFA~\cite{zhang2024bifa} regresses differential flow fields but is unstable under large displacements. ChangeRD~\cite{jing2025changerd} applies adaptive perspective transformations, yet global homographies cannot capture local geometric variation. SimSaC~\cite{park2022simsac} jointly predicts flow and change masks using synthetic cut-paste supervision, but its artificial perturbations poorly reflect real temporal variation. These approaches illustrate the promise of coupling registration and detection, but remain limited by implicit alignment, regression instability, or unrealistic training pipelines.  

We propose DiffRegCD, a diffusion-driven framework that explicitly integrates dense registration and change detection. Our approach leverages multi-scale features from a pretrained denoising diffusion model, fine-tuned on large-scale street-view datasets (Pittsburgh 250k ~\cite{Arandjelovic16netvlad}, Tokyo 24/7 ~\cite{7298790}, Tokyo Time Machine ~\cite{rocco2017convolutionalneuralnetworkarchitecture}). These features provide invariance to viewpoint and illumination changes while retaining pixel-level detail. A RoMa-inspired transformer decoder then estimates dense flow from coarse features and propagates alignment across scales. Critically, we reformulate correspondence as a Gaussian-smoothed classification task, which stabilizes training and delivers sub-pixel accuracy under severe misalignment.  

Training is enabled through controlled affine perturbations applied to existing CD datasets (e.g., VL-CMU-CD~\cite{alcantarilla2016street}), yielding paired ground truth for both flow and change labels. This principled strategy avoids pseudo-labels and handcrafted augmentation. Once trained, DiffRegCD generalizes across ground-level and aerial datasets, including VL-CMU-CD, LEVIR-CD, DSIFN-CD, WHU-CD, and SYSU-CD, demonstrating robustness to long temporal gaps and wide viewpoint variation.  

\textbf{Our contributions are as follows:}  
\begin{itemize}
    \item We introduce an integrated framework for registration and change detection, leveraging diffusion-pretrained features for robustness under large temporal gaps.  
    \item We reformulate correspondence as Gaussian-smoothed classification, achieving stable optimization and sub-pixel accuracy beyond regression-based methods.  
    \item We design a supervised pipeline using synthetic affine perturbations, providing paired ground truth for both flow and change detection.  
    \item We establish state-of-the-art results on both ground-level (VL-CMU-CD) and remote sensing (LEVIR-CD, DSIFN-CD, WHU-CD, SYSU-CD) benchmarks.  
\end{itemize}

\section{Related Works}
\label{sec:related}

\subsection{Image Registration}
Classical registration relied on hand-crafted keypoints such as SIFT~\cite{Lowe:2004}, SURF\cite{Bay2008346}, and ORB\cite{rublee2011orb}, with matches refined via RANSAC\cite{10.1145/358669.358692}. These methods perform well under small viewpoint changes but fail in low-texture regions, repetitive patterns, or large perspective shifts. Learning-based approaches, including SuperPoint~\cite{detone2018superpointselfsupervisedpointdetection} and SuperGlue~\cite{sarlin2020supergluelearningfeaturematching}, improve robustness by jointly detecting and matching keypoints, yet they remain sparse and brittle under occlusion.  

To overcome sparsity, semi-dense models such as LoFTR~\cite{sun2021loftrdetectorfreelocalfeature} and NCNet~\cite{rocco2017convolutionalneuralnetworkarchitecture} match dense features directly, offering broader coverage but still depending heavily on feature distinctiveness. Dense optical flow networks such as RAFT~\cite{teed2020raftrecurrentallpairsfield}, GMFlow~\cite{xu2022gmflowlearningopticalflow}, FlowFormer++~\cite{huang2022flowformertransformerarchitectureoptical}, and RoMa~\cite{edstedt2023romarobustdensefeature} produce pixel-wise correspondences suitable for alignment. However, most rely on photometric consistency, which breaks under illumination changes, occlusion, or long temporal gaps.

\subsection{Change Detection under Co-Registration}
Many CD methods assume perfectly aligned inputs. Early CNNs such as FC-EF~\cite{daudt2018fullyconvolutionalsiamesenetworks} and FC-Siam~\cite{bertinetto2021fullyconvolutionalsiamesenetworksobject} fused features across time, while more advanced designs introduced transformers for global reasoning, e.g., BIT-CD~\cite{Chen_2022} and ChangeFormer~\cite{bandara2022transformerbasedsiamesenetworkchange}. More recent methods leverage large pretrained backbones, including DDPM-CD~\cite{bandara2022ddpm}, to improve representation quality. These approaches achieve strong results on curated datasets such as LEVIR-CD, WHU-CD, DSIFN-CD, and SYSU-CD, but they degrade sharply once inputs are misaligned: even minor shifts cause false positives to dominate.

\subsection{Joint Registration and Change Detection}
Several works have attempted to couple registration and detection. SimSaC~\cite{park2022simsac} jointly estimates scene flow and change masks using synthetic cut–paste perturbations, though such supervision fails to capture real temporal variation. BiFA~\cite{zhang2024bifa} introduces bidirectional feature alignment via flow regression, but suffers instability under large displacements. ChangeRD~\cite{jing2025changerd} integrates adaptive perspective transforms with attention-guided convolutions to suppress pseudo-changes, yet its reliance on global homographies limits local precision. URCNet~\cite{10366319} incorporates affine modules for remote sensing CD, but struggles with fine-scale distortions. While these frameworks highlight the promise of joint modeling, they often rely on regression-only flow, global transformations, or synthetic supervision that does not generalize across domains.

\subsection{Discussion}


Existing registration methods, from sparse keypoints to dense transformers, degrade under long temporal baselines, while CD models perform well only under accurate co-registration and fail under misalignment. Joint frameworks partially address this but remain limited by implicit alignment or unstable supervision. Our approach explicitly unifies dense registration and change detection. DiffRegCD leverages diffusion-pretrained, invariant features, formulates correspondence as classification for sub-pixel alignment, and applies supervised multi-scale warping with paired flow and change labels, enabling robust CD under severe misalignment.

\begin{figure*}[t]
  \centering
  \includegraphics[width=\textwidth]{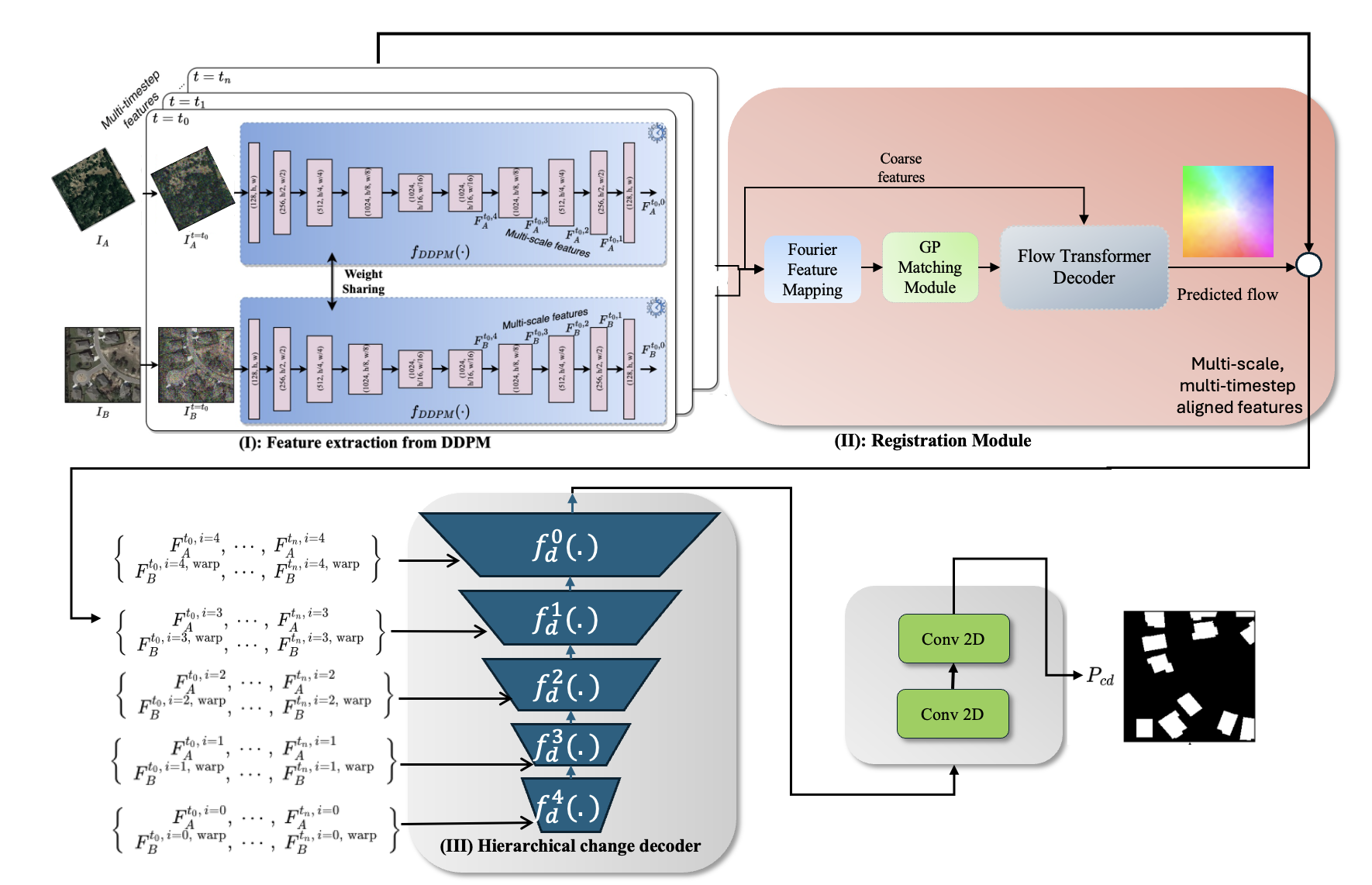}
  \caption{\textbf{Pipeline of our proposed framework.} 
  (I) A DDPM-based encoder extracts multi-scale, multi-timestep features from the bi-temporal inputs $I_A$ and $I_B$. 
  (II) A registration module, consisting of a Gaussian Prior and a Flow Transformer Decoder, aligns features with coordinate embeddings to estimate dense flows. 
  (III) A hierarchical change decoder fuses the warped multi-scale features across resolutions to predict the final change map $P_{cd}$. 
  This design provides robustness to misregistration while leveraging diffusion-based features for accurate change detection. \emph{The overall visualization style is inspired by the pipeline illustration in the DDPM-CD model.}}
  \label{fig:pipeline_overview}
\end{figure*}

\section{Methodology}
\label{sec:method}

\noindent\textbf{Problem definition.}
Given a pre/post image pair $I^A,I^B\!\in\!\mathbb{R}^{3\times H\times W}$, we seek (i) a dense displacement field $W_{B\!\to\!A}\!\in\!\mathbb{R}^{2\times H\times W}$ aligning $I^B$ to $I^A$, and (ii) a change mask $M\!\in\!\{0,1\}^{H\times W}$. Pixel coordinates are $\Omega=\{1{:}H\}\!\times\!\{1{:}W\}$. For $x\!\in\!\Omega$, vectors are bold, $\|\cdot\|_2$ is the $\ell_2$ norm, and $\odot$ denotes elementwise product. Bilinear warping of feature $f$ by displacement $U$ is
\[
\mathcal{W}(f,U)(x)=\!\!\sum_{p\in\mathcal{N}(x+U(x))}\!\!\kappa(x{+}U(x),p)f(p),
\]
where $\kappa$ is the bilinear kernel and $\mathcal{N}(\cdot)$ its four neighbors.

\subsection{Diffusion-Pretrained Features}
We employ a frozen denoising diffusion encoder $\phi$, trained on large-scale natural imagery, as a universal feature extractor. For an input $I$, $\phi$ provides a hierarchy of features indexed by the diffusion timestep $t \in \mathcal{T}$ and the pyramid scale $i \in \{0,\dots,4\}$, corresponding to $L=4$ encoder levels spanning resolutions from full scale down to $1/16$:

\[
\mathcal{F}(I)=\{\,f^{t,i}\in\mathbb{R}^{C_i\times H_i\times W_i}\,\},\qquad H_i=H/2^i.
\]

Small $t$ correspond to lightly noised inputs and thus preserve high-frequency edges and textures; large $t$ are closer to semantic denoising stages, encoding shape and category information invariant to illumination, season, or sensor modality. Here, ``large'' timesteps refer to late-stage diffusive steps with high noise levels, and we select a large but not necessarily the final timestep in practice.
 This duality provides both local detail and global invariance, which are critical for misaligned change detection. The spatial pyramid captures features from coarse (contextual, semantic) to fine (local, geometric) levels. Together with timestep variation, this yields a two-dimensional grid of features $\{f^{t,i}\}$ spanning both resolution and semantic abstraction. Since $C_i$ varies with scale and timestep, each feature map is projected to a fixed dimension $D_i$ with a $1{\times}1$ linear map $P_i$ followed by LayerNorm:
\[
\bar f^{t,i}=\mathrm{LN}(P_i f^{t,i}),
\]
ensuring comparability across scales/timesteps and stabilizing downstream attention. We use bar notation ($\bar f^{t,i}$) for harmonized features throughout.

Coarse semantic features ($t$ large, $i=L$) provide stable descriptors for registration, while the full multi-scale set $\{\bar f^{t,i}\}$ is reserved for the change detection head. Importantly, $\phi$ remains frozen to preserve invariances learned during diffusion pretraining, and only the projection layers and subsequent modules are trained.

\subsection{Gaussian-Processed Transformer for Flow}
\label{sec:flow}

\noindent\textbf{Fourier GP matching.}
We estimate flow from the coarse semantic DDPM~\cite{ho2020denoisingdiffusionprobabilisticmodels} feature at decoder block $i^\star{=}11$ and timestep $t^\star$, where $i^\star$ denotes the depth within the diffusion U\mbox{-}Net and is distinct from the multi\mbox{-}scale pyramid index.
Before the transformer, the coarse features $(\bar f_A^{t^\star,11}, \bar f_B^{t^\star,11})$ are processed by a GP-style matching module with a cosine kernel and Fourier feature mapping,
\[
k(f_i,f_j) \approx \phi(f_i)^\top \phi(f_j), 
\qquad 
\phi(f) = [\cos(Wf), \sin(Wf)].
\]
This module produces a displacement-wise correlation volume that serves as input to the flow transformer decoder and induces a smooth GP-like prior over the predicted flow.
\noindent\textbf{Transformer decoder.}  
At the selected coarse scale, we concatenate the GP posterior and image features and reshape them into a sequence of spatial tokens with positional encoding:
\[
T = \mathrm{reshape}([\mathrm{GP}, f]) + \pi, \quad T \in \mathbb{R}^{B \times HW \times C}.
\]
The tokens are processed by a stack of $S$ transformer blocks with self-attention,
\[
h^{t^\star,11} = \Psi(T),
\]
followed by a linear projection that maps each token to classifier logits and an auxiliary channel. The output is reshaped back to the spatial grid and passed to the registration head for motion decoding.


\noindent\textbf{Registration head.}  
We discretize motion into a single lattice $\mathcal{M}=\{m_k\}_{k=1}^{r^2}$ with spacing $\Delta$ (range $\pm r\Delta$ pixels). A classifier maps $h^{t^\star,11}(x)$ to logits $z(x)$ and a temperature-controlled distribution
\[
\pi(x)=\mathrm{softmax}(z(x)/\tau).
\]
The expected displacement is the \emph{arg-softmax}:
\[
\hat W(x)=\sum_k \pi_k(x)\,m_k.
\]
A lightweight refinement module upsamples $\hat W$ to input resolution, yielding the final flow.

\noindent\textbf{Flow supervision.}  
Ground-truth displacement $w^*(x)$ is projected to lattice scale and encoded as a Gaussian distribution
\[
P(m_k|x)\propto \exp\!\left(-\tfrac{\|m_k-w^*(x)\|^2}{2\sigma^2}\right).
\]
The flow loss combines classification and regression:
\[
\mathcal{L}_{\text{flow}}=\tfrac{1}{|\Omega|}\sum_x 
\Big[\mathrm{KL}(\pi(x)\|P(\cdot|x))+\alpha\|\hat W(x)-w^*(x)\|_2\Big].
\]

The lattice head reframes correspondence as a probabilistic prediction, capturing uncertainty when multiple matches are plausible. The Gaussian target ensures that near-misses are penalized less than distant errors. KL divergence \cite{Kullback1951} trains the classifier to approximate this uncertainty, while the EPE term enforces geometric accuracy. In combination with Gaussian pre-smoothing, this setup provides GP-like regularization: flow remains smooth but adapts to strong image evidence.

\subsection{Aligned Change Detection Head}
Once flow is estimated, we explicitly align post-change features before differencing. For each timestep $t$ and scale $i$, warped features are
\[
\tilde f_B^{t,i}=\mathcal{W}(\bar f_B^{t,i},\hat W),
\]
where $\hat W$ is the full-resolution displacement from Sec.~\ref{sec:flow}. We compute complementary descriptors:
\[
\Delta^{t,i}=|\bar f_A^{t,i}-\tilde f_B^{t,i}|,\qquad
\Pi^{t,i}=\bar f_A^{t,i}\odot \tilde f_B^{t,i},
\]
and concatenate $u^{t,i}=[\Delta^{t,i};\Pi^{t,i}]$. The absolute difference highlights additive discrepancies (direct evidence of change), while the correlation term captures multiplicative agreement (contextual similarity). To reweight informative channels, each $u^{t,i}$ passes through a squeeze–excitation block.

\noindent\textbf{Temporal aggregation.}  
Because DDPM features vary with timestep $t$ (from textural to semantic), we learn attention weights $\gamma^i$ with $\sum_t \gamma_t^i=1$ and form
\[
\hat u^i=\sum_t \gamma_t^i\,\mathrm{SE}(u^{t,i}).
\]

\noindent\textbf{Hierarchical decoder.}  
Aligned representations are decoded coarse-to-fine:
\[
h^i=\varphi_i(\mathrm{Ups}(h^{i+1})\oplus \hat u^i), \quad i=L{-}1,\ldots,0,
\]
where $\varphi_i$ is a light conv block and $\oplus$ denotes concatenation. The final logits $s(x)$ yield per-pixel change probabilities
\[
P_{\text{CD}}(x)=\sigma(s(x)).
\]
Supervision uses binary cross-entropy:
\[
\mathcal{L}_{\text{CD}}=-\tfrac{1}{|\Omega|}\sum_x\Big[Y(x)\log P_{\text{CD}}(x)+(1{-}Y(x))\log(1{-}P_{\text{CD}}(x))\Big].
\]

\subsection{Curriculum and Joint Optimization}
Training both heads jointly is brittle: poor flows at initialization corrupt CD supervision. We therefore adopt a curriculum.  
\emph{Phase I:} train only with $\mathcal{L}_{\text{flow}}$ for $T_{\text{wu}}$ iterations.  
\emph{Phase II:} introduce CD with a ramped weight
\[
\mathcal{L}_{\text{total}}=\mathcal{L}_{\text{flow}}+\lambda_{\text{CD}}(t)\mathcal{L}_{\text{CD}},\quad
\lambda_{\text{CD}}(t)=\lambda_{\max}\min\!\big(1,\tfrac{t-T_{\text{wu}}}{T_{\text{ramp}}}\big).
\]
This schedule ensures alignment precedes semantic differencing, preventing error propagation. Flow-guided warping removes parallax, difference+correlation provide complementary cues, squeeze–excitation emphasizes discriminative channels, and timestep aggregation exploits the DDPM hierarchy. Together with the curriculum, these design choices yield robust change detection under severe misalignment.

\begin{figure*}[t]
  \centering

  \begin{subfigure}[t]{0.49\textwidth}
    \centering
    \includegraphics[width=\linewidth]{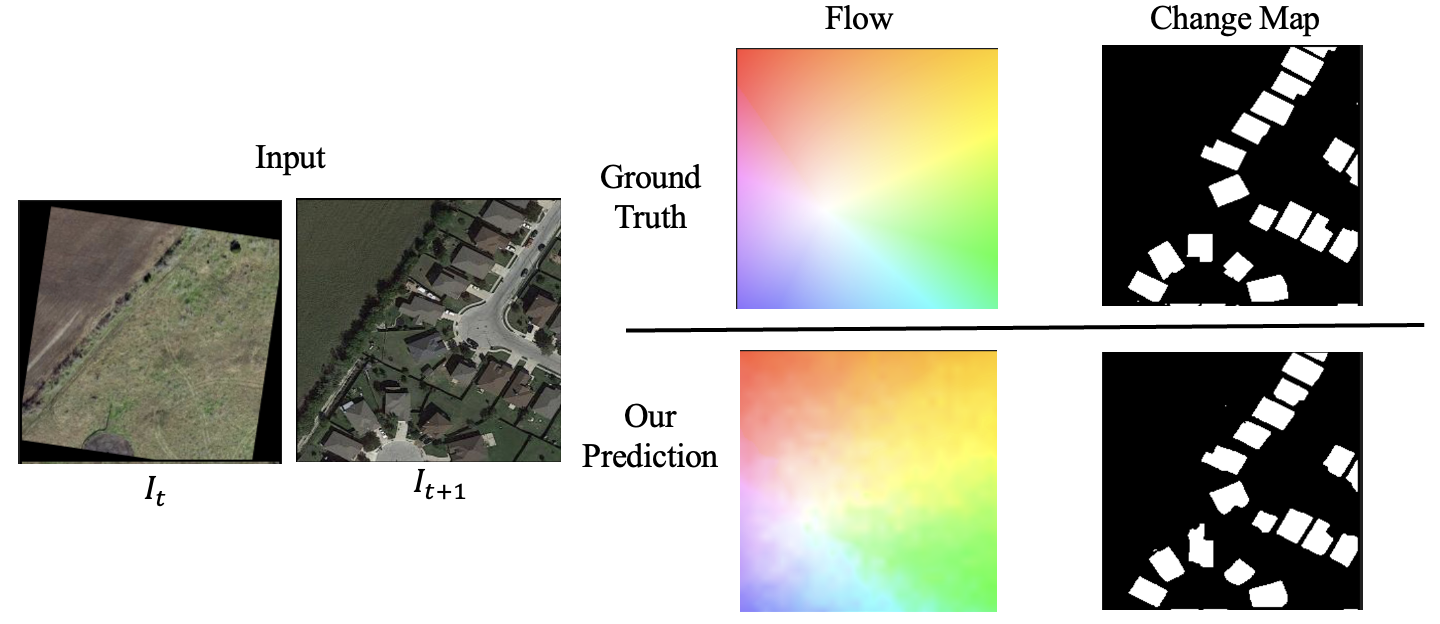}
    \caption{LEVIR-CD}
  \end{subfigure}\hfill
  \begin{subfigure}[t]{0.49\textwidth}
    \centering
    \includegraphics[width=\linewidth]{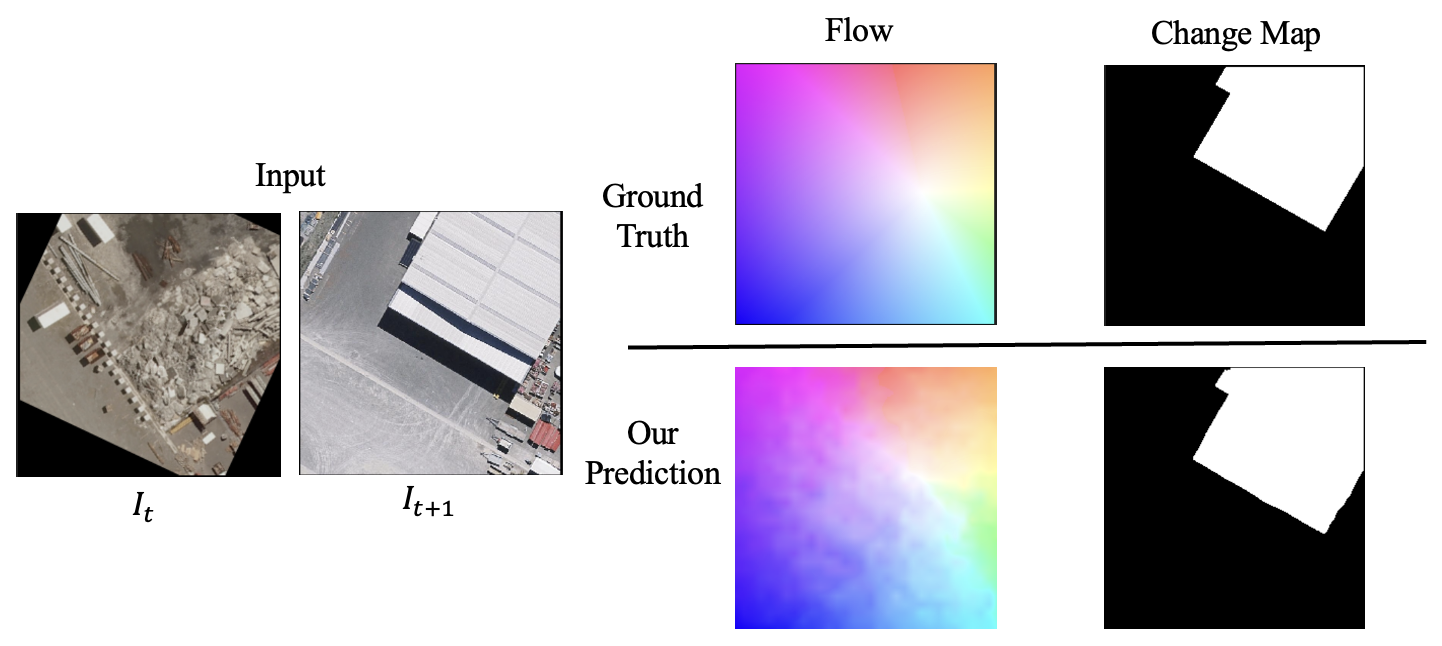}
    \caption{WHU-CD}
  \end{subfigure}\hfill

  \vspace{4pt}

  \begin{subfigure}[t]{0.49\textwidth}
    \centering
    \includegraphics[width=\linewidth]{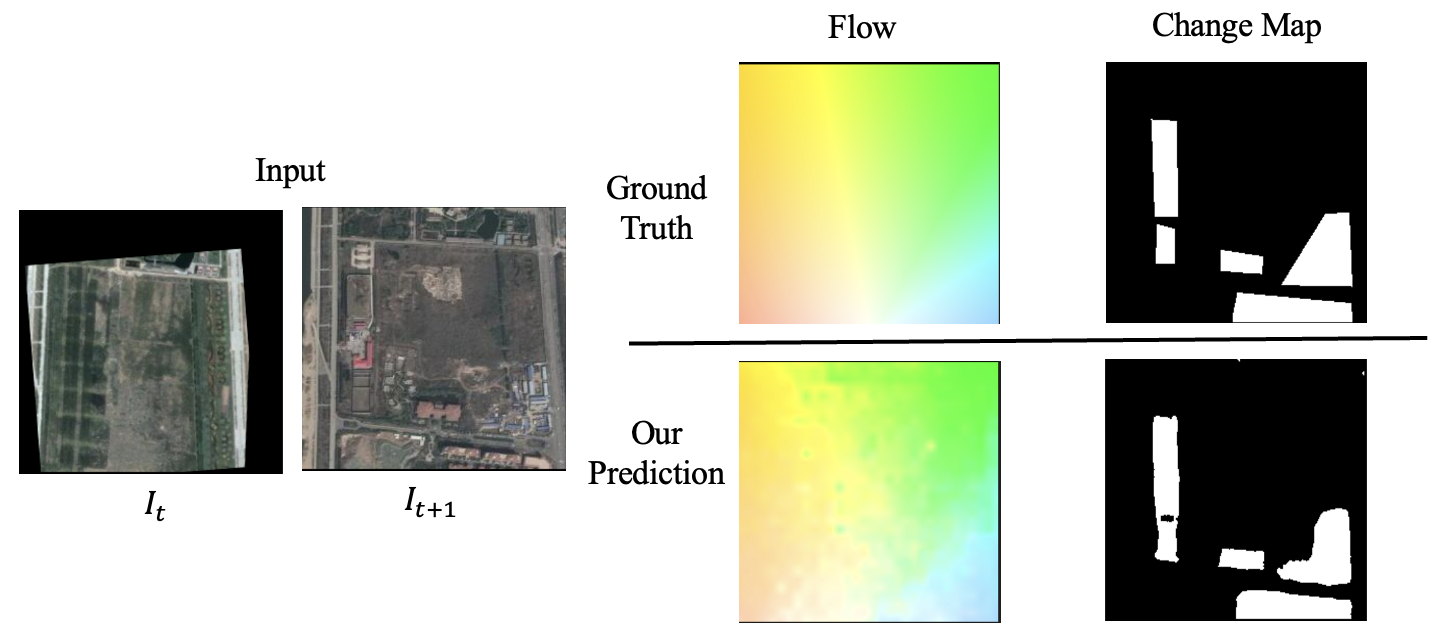}
    \caption{DSIFN-CD}
  \end{subfigure}\hfill
  \begin{subfigure}[t]{0.49\textwidth}
    \centering
    \includegraphics[width=\linewidth]{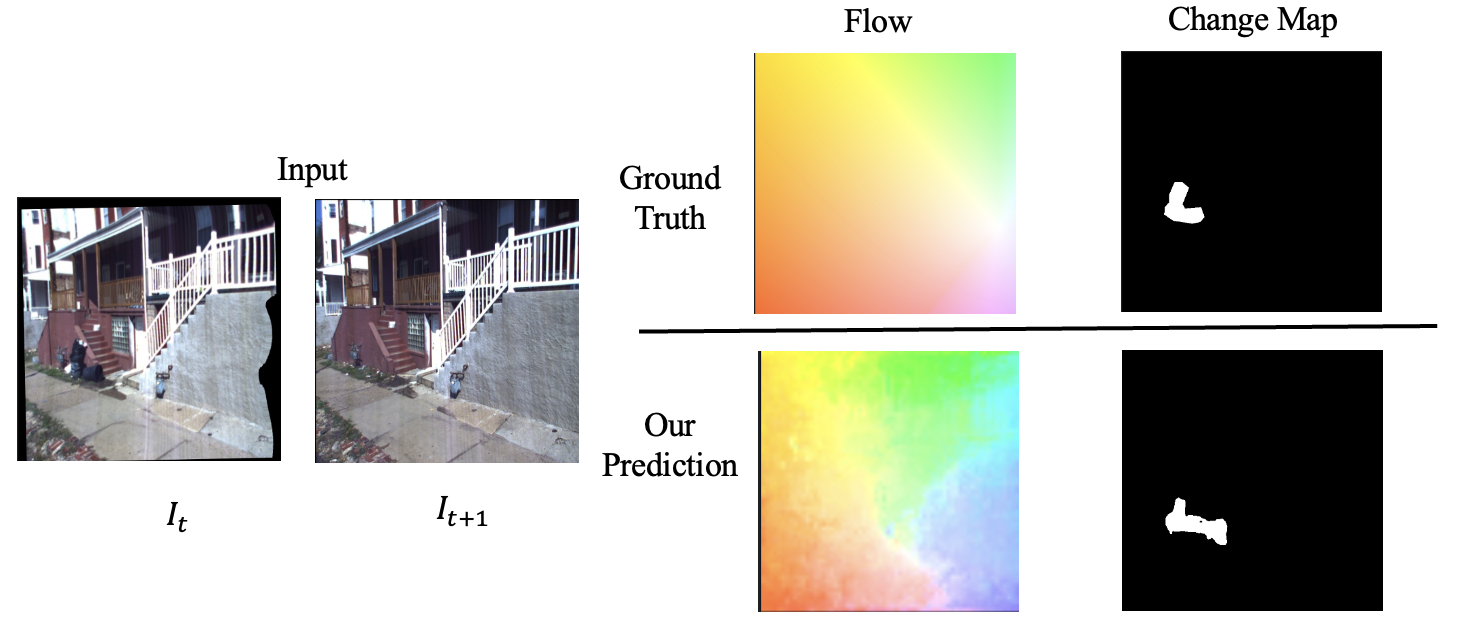}
    \caption{VL-CMU-CD}
  \end{subfigure}
  \caption{\textbf{Cross-dataset qualitative results under induced misregistration.}
  Each mini-panel follows the same layout: left—inputs $I_t$ and $I_{t+1}$; right—top shows ground-truth flow (color-wheel) and change mask; bottom shows our predicted flow and change map. 
  We display diverse aerial and street-level scenes from \emph{LEVIR-CD}, \emph{WHU-CD}, \emph{DSIFN-CD}, \emph{SYSU-CD}, and \emph{VL-CMU-CD}. 
  Patch size $256{\times}256$; level \emph{Hard}: $\Delta x,y\!\in\![-25,25]$ px, $\theta\!\in\![-30^\circ,30^\circ]$, $s\!\in\![0.80,1.25]$. 
  Beyond alignment, our method yields crisp boundaries, fewer background false alarms, and better recovery of small structures across datasets. 
  White denotes change; threshold 0.5; no post-processing.}
  \label{fig:qual_5datasets_readable}
\end{figure*}
\captionsetup[subfigure]{labelformat=parens, labelsep=space, textfont=small}

\section{Experiments}
\label{sec:experiments}


We evaluate DiffRegCD on five representative benchmarks spanning both street-view and satellite imagery: VL-CMU-CD, LEVIR-CD, SYSU-CD, WHU-CD, and DSIFN-CD. These datasets cover diverse spatial resolutions, imaging modalities, and change types. Our evaluation addresses three questions: (i) do diffusion-pretrained features support both dense registration and change detection, (ii) how does the method compare to strong baselines, and (iii) which components are most critical to performance.

\subsection{Datasets}
VL-CMU-CD contains $\sim$260k urban street-view pairs from Pittsburgh with large viewpoint shifts, dynamic objects, and illumination changes. We follow the standard 250k/10k split and resize to $256^2$.  
LEVIR-CD consists of 637 pairs of 1024$\times$1024 satellite images with building-level changes collected over multiple years. Following prior work, images are cropped into $256^2$ patches, yielding $\sim$31k training and 10k test pairs.  
SYSU-CD provides 20k multispectral satellite pairs at 256$\times$256 resolution with heterogeneous land-cover changes (urban, farmland, vegetation).  
WHU-CD contains 20 aerial image pairs of 3250$\times$1530 resolution with building annotations, producing $\sim$20k 256$\times$256 patches. Seasonal differences and high scene variability make alignment difficult.  
DSIFN-CD focuses on fine-grained object changes with 4000 pairs of 512$\times$512 images cropped into 36k 256$\times$256 patches. Subtle local differences and radiometric distortions present strong challenges.  

For all datasets, we adopt official train/val/test splits. Change detection supervision uses the provided binary masks. Registration supervision during training is derived from synthetic affine/projective transformations with analytically computed flows (see Sec.~\ref{sec:experiments}). Unless otherwise specified, results are reported as mean$\pm$std over three random seeds.\\
\noindent\textbf{Data splits and supervision.}
For all datasets, we adopt the official train/val/test splits. Change-detection supervision uses the provided binary masks. Registration supervision during training is obtained from synthetic affine perturbations with analytically computed dense flows (forward warps from the sampled transform); the default sampling ranges are listed in Table~\ref{tab:proto_affine_default}. Unless otherwise specified, results are reported as mean$\pm$std over three seeds.
\vspace{-5mm}

\begin{figure*}[t]
  \centering
  \includegraphics[width=\textwidth]{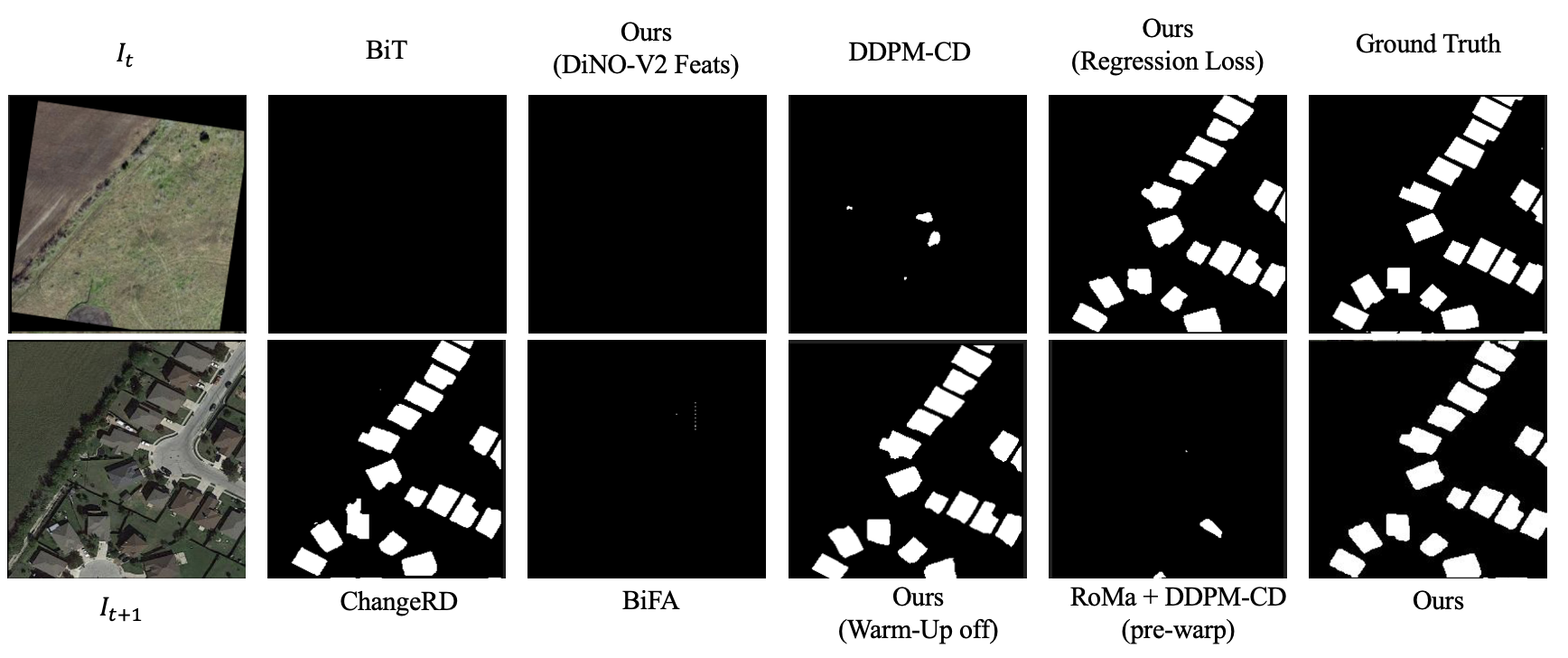} 
  \caption{\textbf{Qualitative comparison under induced misalignment.}
  Dataset: LEVIR-CD, patch size $256{\times}256$. 
  Level \emph{Hard}: $\Delta x,y\!\in\![-25,25]$ px, $\theta\!\in\![-30^{\circ},30^{\circ}]$, $s\!\in\![0.80,1.25]$.
  Columns show inputs $I_t$ and $I_{t+1}$, predictions from \textit{BiT}, \textit{ChangeRD}, \textit{BiFA}, \textit{DDPM-CD}, \textit{RoMa$\!\to$CD}, ablations of our method, our full model, and the ground-truth mask (white = change).
  All methods use the same threshold (0.5) and no post-processing.}
  \label{fig:qual_levircd_hard}
\end{figure*}

\paragraph{Baselines and metrics.}
We compare against three categories of methods:  
(i) \emph{Change-detection only:} ChangeFormer~\cite{bandara2022transformerbasedsiamesenetworkchange}, BIT~\cite{Chen_2022}, and BiFA~\cite{zhang2024bifa}, which represent state-of-the-art transformer and feature-aggregation approaches but do not explicitly address misalignment.  
(ii) \emph{Unified registration+CD:} ChangeRD~\cite{jing2025changerd}, which integrates spatial alignment and change prediction in a single network.  
(iii) \emph{Registration only:} SuperPoint+SuperGlue~\cite{sarlin2020supergluelearningfeaturematching}, LoFTR~\cite{sun2021loftrdetectorfreelocalfeature}, RoMa~\cite{edstedt2023romarobustdensefeature}, and MASt3R~\cite{leroy2024groundingimagematching3d}, evaluated by applying their estimated correspondences to align pairs prior to CD.  
Other joint models (SimSaC~\cite{park2022simsac}, URCNet~\cite{zhou2023unified}) were not included due to incomplete or unavailable implementations.  

For \emph{change detection}, we report F1-score, mean Intersection over Union (mIoU), and Overall Accuracy (OA), which together capture boundary quality, region consistency, and global classification correctness.  

Hyperparameters and training protocols are kept consistent across baselines to ensure fair comparison.

\begin{table}[h]
\centering
\footnotesize
\resizebox{\columnwidth}{!}{ \begin{tabular}{lcccc}
\toprule
\textbf{Level} & $\Delta x,y$ [px] & $\theta$ [°] & $s$ (×) & Train/Test \\
\midrule
Default & U[-25,25] & U[-30,30] & U[0.80,1.25] & \checkmark / \checkmark \\
\bottomrule
\end{tabular}}
\caption{\textbf{Synthetic perturbation protocol (affine-only).} Ranges match the script: random translation $(\Delta x,\Delta y)$, rotation $\theta$, and scale $s$.}
\label{tab:proto_affine_default}
\end{table}

\noindent\textbf{Implementation and training details.} 
All models are implemented in \textbf{PyTorch} and trained on a single NVIDIA \textbf{A6000 GPU} with 48 GB of memory. Unless otherwise noted, we use the \textbf{AdamW} optimizer with an initial learning rate of $1 \times 10^{-4}$, cosine annealing schedule, and gradient accumulation over 8 steps. A warm-up phase of 500 iterations is applied to stabilize optimization; ablations without warm-up are reported separately. The batch size is fixed at 2 due to memory constraints, and all input images are resized to $256 \times 256$. Training typically converges within 100 epochs. To mitigate randomness, each experiment is repeated three times with different seeds, and average results are reported.
We also report model complexity, parameter counts (M) and FLOPs (G at $256{\times}256$), for each component and baseline comparison in the supplementary material.

\noindent\textbf{Evaluation metrics.} 
Performance is assessed using four complementary metrics: \textbf{mean F1 (mF1)}, \textbf{mean IoU (mIoU)}, \textbf{overall accuracy (OA)}, and \textbf{class-specific F1 for the change class (F1$_1$)}. OA captures global pixel-level correctness, mIoU measures region overlap, mF1 balances precision and recall across both classes, and F1$_1$ specifically quantifies accuracy on the change class, the most critical aspect for practical CD applications.

\begin{table}[t]
\centering
\scriptsize
\resizebox{\columnwidth}{!}{ \begin{tabular}{lcccc}
\toprule
\textbf{LEVIR-CD} & \textbf{mF1} & \textbf{mIoU} & \textbf{OA} & \textbf{F1$_1$} \\
\midrule
\multicolumn{5}{l}{\emph{Baselines}} \\
BiFA \cite{zhang2024bifa} & 0.490 & 0.475 & 0.947 & 0.006 \\
ChangeFormer \cite{bandara2022transformerbasedsiamesenetworkchange} & 0.914 & 0.852 & 0.985 & 0.8489\\
BIT-CD  \cite{Chen_2022}      & 0.915 & 0.853 & 0.985 & 0.8385 \\
DDPM-CD    \cite{bandara2022ddpm}   & 0.916 & 0.856 & 0.986 & 0.8397 \\
ChangeRD   \cite{jing2025changerd}              & 0.929 & 0.875  & 0.980 & 0.866\\
\addlinespace[2pt]
\textbf{Ours}            & \textbf{0.929} & \textbf{0.881} & \textbf{0.987} &  \textbf{0.872} \\
\bottomrule
\end{tabular}}
\caption{Comparison of baseline CD backbones and ChangeRD with our approach on the LEVIR-CD dataset. All methods are trained on the synthetic dataset and evaluated on LEVIR-CD, reporting mF1, mIoU, OA, and F1$_1$.}
\label{tab:levir_results}
\end{table}

\subsection{Quantitative Results}

Tables ~\ref{tab:levir_results}-~\ref{tab:sysu_results} summarize performance on LEVIR-CD, WHU-CD, DSIFN-CD, and SYSU-CD. Across all benchmarks, our framework consistently establishes a new state-of-the-art. On WHU-CD, we achieve an mF1 of 0.934 and an F1$_1$ of 0.874, representing a gain of nearly +19 points over the strongest baseline (Table~\ref{tab:whu_results}). On DSIFN-CD, our method reaches an mF1 of 0.940, outperforming DDPM-CD by +5 points (Table~\ref{tab:dsifn_results}). Even on the challenging SYSU-CD dataset, which involves multispectral imagery and large domain gaps, our model delivers an mF1 of 0.910, more than doubling the performance of BiFA (0.430, Table~\ref{tab:sysu_results}). These improvements are consistent across all metrics (mIoU, OA, and F1$_1$), underscoring both the alignment quality and downstream CD accuracy of our approach.

The VL-CMU-CD benchmark explicitly evaluates robustness under induced spatial perturbations of increasing severity. As shown in Table~\ref{tab:vlcmu_results}, our method maintains strong performance across all settings: from 82.1 under low perturbation to 72.0 under high perturbation. In contrast, regression-only approaches such as ChangeRD degrade sharply, dropping to 58.9 under high misalignment. These results highlight that explicitly disentangling and modeling registration yields robustness that conventional CD backbones cannot attain, and confirms the necessity of our unified registration–detection design.

\begin{table}[t]
\vspace{-4mm}
\centering
\scriptsize
\resizebox{\columnwidth}{!}{ \begin{tabular}{lcccc}
\toprule
\textbf{WHU-CD} & \textbf{mF1} & \textbf{mIoU} & \textbf{OA} & \textbf{F1$_1$} \\
\midrule
\multicolumn{5}{l}{\emph{Baselines}} \\
BiFA \cite{zhang2024bifa} & 0.498 & 0.470 & 0.921 & 0.038 \\
BIT-CD  \cite{Chen_2022} & 0.498 & 0.482 & 0.956 & 0.0177 \\
DDPM-CD\cite{bandara2022ddpm}  & 0.745 & 0.647 & 0.945 & 0.5192 \\
ChangeRD  \cite{jing2025changerd}          & 0.705 & 0.608 & 0.930 & 0.447 \\
\addlinespace[2pt]
\textbf{Ours}       & \textbf{0.934} & \textbf{0.883} & \textbf{0.990} & \textbf{0.874} \\
\bottomrule
\end{tabular}}
\caption{Comparison of baseline CD backbones, ChangeRD, and BiFA with our approach on the WHU-CD dataset. All methods are trained on the synthetic dataset and evaluated on WHU-CD, reporting mF1, mIoU, OA, and F1$_1$.}
\label{tab:whu_results}
\end{table}

\begin{table}[t]
\vspace{-4mm}
\centering
\scriptsize
\resizebox{\columnwidth}{!}{ \begin{tabular}{lcccc}
\toprule
\textbf{SYSU-CD} & \textbf{mF1} & \textbf{mIoU} & \textbf{OA} & \textbf{F1$_1$} \\
\midrule
\multicolumn{5}{l}{\emph{Baselines}} \\
BiFA   \cite{zhang2024bifa}            & 0.430 & 0.298 & 0.509 & 0.218 \\
\addlinespace[2pt]
\textbf{Ours}      & \textbf{0.910} & \textbf{0.839} & \textbf{0.941} & \textbf{0.858} \\
\bottomrule
\end{tabular}}
\caption{Comparison of baseline CD backbones and ChangeRD with our approach on the SYSU-CD dataset. All methods are trained on the synthetic dataset and evaluated on SYSU-CD, reporting mF1, mIoU, OA, and F1$_1$.}
\label{tab:sysu_results}
\end{table}

\begin{table}[t]
\vspace{-4mm}
\centering
\scriptsize
\resizebox{\columnwidth}{!}{ \begin{tabular}{lcccc}
\toprule
\textbf{DSIFN-CD} & \textbf{mF1} & \textbf{mIoU} & \textbf{OA} & \textbf{F1$_1$} \\
\midrule
\multicolumn{5}{l}{\emph{Baselines}} \\
BiFA \cite{zhang2024bifa} & 0.519 & 0.391 & 0.652 & 0.266 \\
BIT-CD \cite{Chen_2022}      & 0.449 & 0.406 & 0.809 & 0.004 \\
ChangeRD  \cite{jing2025changerd}   & 0.454 & 0.415 & 0.830 & 0.707 \\
ChangeFormer \cite{bandara2022transformerbasedsiamesenetworkchange} & 0.825 & 0.715 & 0.886 & 0.722 \\

DDPM-CD \cite{bandara2022ddpm}     & 0.889 & 0.806 & 0.930 & 0.840 \\

\addlinespace[2pt]
\textbf{Ours} & \textbf{0.940} & \textbf{0.890} & \textbf{0.967} & \textbf{0.900} \\
\bottomrule
\end{tabular}}
\caption{Comparison of baseline CD backbones, ChangeRD, and BiFA with our approach on the DSIFN-CD dataset. All methods are trained on the synthetic dataset and evaluated on DSIFN-CD, reporting mF1, mIoU, OA, and F1$_1$.}
\label{tab:dsifn_results}
\end{table}

\begin{table}[t]
\vspace{-4mm}
\centering
\scriptsize
\resizebox{\columnwidth}{!}{ \begin{tabular}{lcccc}
\toprule
\textbf{VL-CMU-CD} & \textbf{Acc} & \textbf{mIoU} & \textbf{mF1} & \textbf{F1} \\
\midrule
\multicolumn{5}{l}{\emph{Baselines}} \\
BiFA \cite{zhang2024bifa}      & 0.114 & 0.060 & 0.113 & 0.116 \\
DDPM-CD  \cite{bandara2022ddpm}  & 0.932 & 0.491 & 0.530 & 0.095 \\
ChangeRD \cite{jing2025changerd}  & 0.940 & 0.519 & 0.574 & 0.180 \\
\addlinespace[2pt]
\textbf{Ours} & \textbf{0.942} & \textbf{0.584} & \textbf{0.670} & \textbf{0.370} \\
\bottomrule
\end{tabular}}
\caption{Performance comparison on VL-CMU-CD under induced misalignment. All models were trained on our synthetic dataset and evaluated on VL-CMU-CD. We report overall accuracy (Acc), mean IoU (mIoU), mean F1 (mF1), and change-class F1 (F1$_1$).}
\label{tab:vlcmu_results}
\end{table}

\subsection{Robustness to Induced Misalignment}
We apply synthetic perturbations (translation, rotation, scaling, homography) at three difficulty levels. Table~\ref{tab:vlcmu_results} reports F1 on VL-CMU-CD. Unified baselines degrade sharply at high displacement, while DiffRegCD preserves $>\!80\%$ of aligned performance.

Figures \ref{fig:qual_5datasets_readable} --~\ref{fig:qual_levircd_hard} illustrate how our framework improves both registration and change localization.  

\subsection{Qualitative Results}

\noindent\textbf{Baselines.} As shown in Fig.~\ref{fig:qual_levircd_hard}, DDPM-CD and ChangeFormer leave ghosting artifacts or miss fine details, while our method yields crisp masks aligned with structures. These trends match the $+5$--$19$ point gains in Tables~\ref{tab:levir_results}--\ref{tab:dsifn_results}.  

\noindent\textbf{Robustness.} On VL-CMU-CD (Fig.~\ref{fig:qual_levircd_hard}), ChangeRD deteriorates under perturbations, whereas our model maintains stable boundaries, consistent with Table~\ref{tab:vlcmu_results}.  

\noindent\textbf{Ablations.} Visualizations in Fig.~\ref{fig:qual_5datasets_readable} confirm quantitative results: DINOv2 produces fragmented maps, regression loss blurs flows, and removing warm-up destabilizes training, while our full model preserves clean, accurate predictions.  

\noindent\textbf{Summary.} These results show our approach not only improves metrics but also produces more reliable and visually coherent change maps under diverse conditions.

\begin{table}[h]
\vspace{-5mm}

\centering
\scriptsize
\resizebox{\columnwidth}{!}{ \begin{tabular}{lcccc}
\toprule
\textbf{LEVIR-CD (Ablation Variants)} & \textbf{mF1} & \textbf{mIoU} & \textbf{OA} & \textbf{F1$_1$} \\
\midrule
Regression Loss             & 0.88 & 0.80 & 0.98 & 0.77 \\
No Warmup                   & 0.46 & 0.39 & 0.75 & 0.07 \\
DINO Features               & 0.43 & 0.31 & 0.56 & 0.16 \\
\addlinespace[2pt]
\textbf{Ours}               & \textbf{0.92} & \textbf{0.87} & \textbf{0.98} & \textbf{0.86} \\
\bottomrule
\end{tabular}}
\caption{Ablation study on internal variants of our framework for LEVIR-CD. We evaluate regression-only training, removal of warmup, and DINO feature substitution against our full model. All methods are trained on the synthetic dataset and evaluated on LEVIR-CD, reporting mF1, mIoU, OA, and F1$_1$.}
\label{tab:levir_ablation_variants}
\vspace{-10pt}
\end{table}

\begin{table}[h]
\centering
\scriptsize
\resizebox{\columnwidth}{!}{ \begin{tabular}{lcccc}
\toprule
\textbf{LEVIR-CD (External Baselines)} & \textbf{mF1} & \textbf{mIoU} & \textbf{OA} & \textbf{F1$_1$} \\
\midrule
RoMa \cite{edstedt2023romarobustdensefeature}  & 0.57 & 0.51 & 0.93 & 0.18 \\
LoFTR \cite{sun2021loftrdetectorfreelocalfeature} & 0.54 & 0.50 & 0.94 & 0.12 \\
MASt3R \cite{leroy2024groundingimagematching3d}& 0.50 & 0.48 & 0.94 & 0.03 \\
\addlinespace[2pt]
\textbf{Ours}               & \textbf{0.92} & \textbf{0.87} & \textbf{0.98} & \textbf{0.86} \\
\bottomrule
\end{tabular}}
\caption{Comparison with external baselines on the LEVIR-CD dataset. We report results for DDPM-CD+RoMa, DDPM-CD+LoFTR, and DDPM-CD+MASt3R against our proposed model. All methods are trained on the synthetic dataset and evaluated on LEVIR-CD, reporting mF1, mIoU, OA, and F1$_1$.}
\label{tab:levir_external}
\end{table}

\subsection{Ablation Studies}
\label{sec:ablation}

We conduct ablations on \textbf{LEVIR-CD} to dissect the contribution of individual components in our framework (Tables~\ref{tab:levir_external} and \ref{tab:levir_ablation_variants}). Each variant is trained under identical settings as our main model (synthetic perturbation dataset, $256 \times 256$ inputs, A6000 GPU) to ensure comparability.

\noindent\textbf{External matchers.} 
We first substitute our registration module with generic correspondence models: (i) DDPM-CD combined with RoMa (using pretrained weights), (ii) LoFTR, and (iii) MASt3R. All three baselines collapse to mF1 $\leq$ 0.572 (Table~\ref{tab:levir_external}), far below our 0.929. This confirms that existing matchers—though effective in isolated tasks—cannot directly integrate into the CD pipeline. They either overfit to natural image domains (LoFTR, MASt3R) or fail to handle the structured misalignments in remote sensing (RoMa pretrained).

\noindent\textbf{Regression vs.\ classification loss.} 
Our model predicts dense flow by formulating displacement estimation as a \emph{classification problem}, where Gaussian smoothing distributes probability mass around the true correspondence. Replacing this with a conventional $L_2$ regression objective (Regression Loss in Table~\ref{tab:levir_ablation_variants}) reduces performance to mF1 = 0.884. Regression collapses when flow distributions are multimodal (e.g., repetitive building facades), whereas classification preserves sharper probability peaks, yielding more stable downstream CD.

\noindent\textbf{Warm-up scheduling.} 
We adopt a 500-iteration warm-up schedule to gradually balance registration and CD losses. Removing warm-up destabilizes optimization, causing gradients from the CD head to dominate early training. As shown in Table~\ref{tab:levir_ablation_variants}, this variant drops catastrophically to mF1 = 0.463. Warm-up thus plays a crucial role in stabilizing joint training of registration and CD modules.

\noindent\textbf{Feature backbone: DINOv2\cite{oquab2023dinov2} vs.\ DDPM \cite{ho2020denoisingdiffusionprobabilisticmodels}.} 
We replace our diffusion-based encoder with a strong self-supervised alternative, DINOv2. Despite its success in recognition tasks, the DINOv2 variant yields only mF1 = 0.436 (Table~\ref{tab:levir_ablation_variants}). This gap arises because DDPM features encode \emph{multi-timestep representations}, capturing progressive noise-to-signal transitions that enrich geometric cues for flow estimation. In contrast, DINOv2 produces temporally invariant embeddings that lack diversity across scales, leading to degraded flow fields and noisier change maps. Qualitative results in Figure~\ref{fig:qual_levircd_hard} illustrate this failure: the DINOv2 backbone misaligns building facades and generates spurious detections.

The ablations validate three key insights: (i) diffusion-pretrained features are critical for robust alignment, (ii) Gaussian classification loss outperforms regression in multimodal correspondence settings, and (iii) warm-up scheduling is indispensable for stable optimization. These design choices enable our framework to consistently outperform external matchers and ablated baselines.
\vspace{-5mm}

\section{Conclusion}
\label{sec:conclusion}

We introduced a unified framework that jointly addresses dense image registration and change detection by leveraging multi-scale features from pretrained Denoising Diffusion Probabilistic Models (DDPMs) \cite{ho2020denoisingdiffusionprobabilisticmodels}. Our design couples diffusion-derived representations with a transformer-based registration decoder and a hierarchical CD head, enabling robust alignment and accurate differencing within a single end-to-end model. To the best of our knowledge, this is the first work to explicitly unify registration and CD using diffusion-pretrained features. Comprehensive experiments on five benchmarks, spanning street-view and remote sensing domains, demonstrate clear gains over prior unified approaches such as ChangeRD \cite{jing2025changerd} and SimSaC \cite{park2022simsac}. Our method consistently delivers higher mF1 and F1$_1$ scores while also reducing flow error, confirming that diffusion features (originally trained for generative modeling) form a powerful basis for geometric alignment under misregistration.


\textbf{Limitations and Future Work.} The current framework remains memory-intensive due to multi-scale aggregation and does not explicitly model flow uncertainty. Future work includes adaptive resolutions, uncertainty-aware decoding, and incorporating additional modalities (e.g., LiDAR, hyperspectral) to enhance long-term real-world robustness.

\section*{Acknowledgements}
This work was supported by the NSF CAREER award under Grant 2045489.

{
    \small
    \bibliographystyle{ieeenat_fullname}
    \bibliography{main}

@String(CVPR= {IEEE Conf. Comput. Vis. Pattern Recog.})

@String(CVPR  = {CVPR})

@article{ji2018fully,
  title={Fully Convolutional Networks for Multi-Source Building Extraction from An Open Aerial and Satellite Imagery Dataset},
  author={Ji, Shunping and Wei, Shiqing and Lu, Meng},
  journal={IEEE Transactions on Geoscience and Remote Sensing},
  year={2018},
  volume={56},
  number={10},
  pages={5743--5756},
  publisher={IEEE},
  doi={10.1109/TGRS.2018.2858817}
}

@misc{oquab2023dinov2,
  title={DINOv2: Learning Robust Visual Features without Supervision},
  author={Oquab, Maxime and Darcet, Timothée and Moutakanni, Theo and Vo, Huy V. and Szafraniec, Marc and Khalidov, Vasil and Fernandez, Pierre and Haziza, Daniel and Massa, Francisco and El-Nouby, Alaaeldin and Howes, Russell and Huang, Po-Yao and Xu, Hu and Sharma, Vasu and Li, Shang-Wen and Galuba, Wojciech and Rabbat, Mike and Assran, Mido and Ballas, Nicolas and Synnaeve, Gabriel and Misra, Ishan and Jegou, Herve and Mairal, Julien and Labatut, Patrick and Joulin, Armand and Bojanowski, Piotr},
  journal={arXiv:2304.07193},
  year={2023}
}

@article{ZHANG2020183,
title = {A deeply supervised image fusion network for change detection in high resolution bi-temporal remote sensing images},
journal = {ISPRS Journal of Photogrammetry and Remote Sensing},
volume = {166},
pages = {183-200},
year = {2020},
issn = {0924-2716},
doi = {https://doi.org/10.1016/j.isprsjprs.2020.06.003},
url = {https://www.sciencedirect.com/science/article/pii/S0924271620301532},
author = {Chenxiao Zhang and Peng Yue and Deodato Tapete and Liangcun Jiang and Boyi Shangguan and Li Huang and Guangchao Liu}
}

@inproceedings{rublee2011orb,
  title={ORB: an efficient alternative to SIFT or SURF},
  author={Rublee, Ethan and Rabaud, Vincent and Konolige, Kurt and Bradski, Gary R},
  booktitle={2011 International conference on computer vision},
  pages={2564--2571},
  year={2011},
  organization={Ieee}
}

@article{Kullback1951,
  author    = {Kullback, Solomon and Leibler, Richard A.},
  title     = {On Information and Sufficiency},
  journal   = {The Annals of Mathematical Statistics},
  volume    = {22},
  number    = {1},
  pages     = {79--86},
  year      = {1951},
  publisher = {Institute of Mathematical Statistics},
  doi       = {10.1214/aoms/1177729694},
  url       = {https://projecteuclid.org/journals/annals-of-mathematical-statistics/volume-22/issue-1/On-Information-and-Sufficiency/10.1214/aoms/1177729694.full}
}

@article{10.1145/358669.358692,
author = {Fischler, Martin A. and Bolles, Robert C.},
title = {Random sample consensus: a paradigm for model fitting with applications to image analysis and automated cartography},
year = {1981},
issue_date = {June 1981},
publisher = {Association for Computing Machinery},
address = {New York, NY, USA},
volume = {24},
number = {6},
issn = {0001-0782},
url = {https://doi.org/10.1145/358669.358692},
doi = {10.1145/358669.358692},
journal = {Commun. ACM},
month = jun,
pages = {381–395},
numpages = {15}
}

@article{Bay2008346,
  author = {Bay, Herbert and Ess, Andreas and Tuytelaars, Tinne and Gool, Luc Van},
  description = {ScienceDirect.com - Computer Vision and Image Understanding - Speeded-Up Robust Features (SURF)},
  doi = {10.1016/j.cviu.2007.09.014},
  issn = {1077-3142},
  journal = {Computer Vision and Image Understanding},
  note = {Similarity Matching in Computer Vision and Multimedia},
  number = 3,
  pages = {346 - 359},
  title = {Speeded-Up Robust Features (SURF)},
  url = {http://www.sciencedirect.com/science/article/pii/S1077314207001555},
  volume = 110,
  year = 2008
}

@misc{ho2020denoisingdiffusionprobabilisticmodels,
      title={Denoising Diffusion Probabilistic Models}, 
      author={Jonathan Ho and Ajay Jain and Pieter Abbeel},
      year={2020},
      eprint={2006.11239},
      archivePrefix={arXiv},
      primaryClass={cs.LG},
      url={https://arxiv.org/abs/2006.11239}, 
}

@ARTICLE{shi21deeply,
  author={Shi, Qian and Liu, Mengxi and Li, Shengchen and Liu, Xiaoping and Wang, Fei and Zhang, Liangpei},
  journal={IEEE Transactions on Geoscience and Remote Sensing}, 
  title={A Deeply Supervised Attention Metric-Based Network and an Open Aerial Image Dataset for Remote Sensing Change Detection}, 
  year={2021},
  volume={},
  number={},
  pages={1-16},
  doi={10.1109/TGRS.2021.3085870}}

@article{jing2025changerd,
  title={ChangeRD: A registration-integrated change detection framework for unaligned remote sensing images},
  author={Jing, Wei and Chi, Kaichen and Li, Qiang and Wang, Qi},
  journal={ISPRS Journal of Photogrammetry and Remote Sensing},
  volume={220},
  pages={64--74},
  year={2025},
  publisher={Elsevier}
}

@article{Arandjelovic16netvlad,
  author    = {Arandjelović, Relja and Gronát, Petr and Torii, Akihiko and Sivic, Josef and Pajdla, Tomáš},
  title     = {NetVLAD: CNN Architecture for Weakly Supervised Place Recognition},
  journal   = {IEEE Transactions on Pattern Analysis and Machine Intelligence},
  year      = {2018},
  volume    = {40},
  number    = {6},
  pages     = {1437--1452},
  publisher = {IEEE},
  doi       = {10.1109/TPAMI.2017.2757247},
  eprint    = {1511.07247},
  archiveprefix = {arXiv},
  primaryclass = {cs.CV}
}

@misc{rocco2017convolutionalneuralnetworkarchitecture,
      title={Convolutional neural network architecture for geometric matching}, 
      author={Ignacio Rocco and Relja Arandjelović and Josef Sivic},
      year={2017},
      eprint={1703.05593},
      archivePrefix={arXiv},
      primaryClass={cs.CV},
      url={https://arxiv.org/abs/1703.05593}, 
}

@inproceedings{alcantarilla2016street,
  title={Street-view change detection with deconvolutional networks},
  author={Alcantarilla, Pablo F and Stent, Simon and Stretton, David and Stamos, Iasonas},
  booktitle={Robotics: Science and Systems (RSS)},
  year={2016}
}

@Article{Lowe:2004,
  author  = {Lowe, David G.},
  title   = {Distinctive Image Features from Scale-Invariant Keypoints},
  journal = {International Journal of Computer Vision},
  year    = {2004},
  volume  = {60},
  number  = {2},
  pages   = {91--110},
  url     = {http://www.cs.ubc.ca/~lowe/papers/ijcv04.pdf}
}

@misc{detone2018superpointselfsupervisedpointdetection,
      title={SuperPoint: Self-Supervised Interest Point Detection and Description}, 
      author={Daniel DeTone and Tomasz Malisiewicz and Andrew Rabinovich},
      year={2018},
      eprint={1712.07629},
      archivePrefix={arXiv},
      primaryClass={cs.CV},
      url={https://arxiv.org/abs/1712.07629}, 
}

@misc{sarlin2020supergluelearningfeaturematching,
      title={SuperGlue: Learning Feature Matching with Graph Neural Networks}, 
      author={Paul-Edouard Sarlin and Daniel DeTone and Tomasz Malisiewicz and Andrew Rabinovich},
      year={2020},
      eprint={1911.11763},
      archivePrefix={arXiv},
      primaryClass={cs.CV},
      url={https://arxiv.org/abs/1911.11763}, 
}

@article{Chen_2022,
   title={Remote Sensing Image Change Detection With Transformers},
   volume={60},
   ISSN={1558-0644},
   url={http://dx.doi.org/10.1109/TGRS.2021.3095166},
   DOI={10.1109/tgrs.2021.3095166},
   journal={IEEE Transactions on Geoscience and Remote Sensing},
   publisher={Institute of Electrical and Electronics Engineers (IEEE)},
   author={Chen, Hao and Qi, Zipeng and Shi, Zhenwei},
   year={2022},
   pages={1–14} }

@misc{daudt2018fullyconvolutionalsiamesenetworks,
      title={Fully Convolutional Siamese Networks for Change Detection}, 
      author={Rodrigo Caye Daudt and Bertrand Le Saux and Alexandre Boulch},
      year={2018},
      eprint={1810.08462},
      archivePrefix={arXiv},
      primaryClass={cs.CV},
      url={https://arxiv.org/abs/1810.08462}, 
}

@misc{leroy2024groundingimagematching3d,
      title={Grounding Image Matching in 3D with MASt3R}, 
      author={Vincent Leroy and Yohann Cabon and Jérôme Revaud},
      year={2024},
      eprint={2406.09756},
      archivePrefix={arXiv},
      primaryClass={cs.CV},
      url={https://arxiv.org/abs/2406.09756}, 
}

@misc{bertinetto2021fullyconvolutionalsiamesenetworksobject,
      title={Fully-Convolutional Siamese Networks for Object Tracking}, 
      author={Luca Bertinetto and Jack Valmadre and João F. Henriques and Andrea Vedaldi and Philip H. S. Torr},
      year={2021},
      eprint={1606.09549},
      archivePrefix={arXiv},
      primaryClass={cs.CV},
      url={https://arxiv.org/abs/1606.09549}, 
}

@INPROCEEDINGS{7298790,
  author={Torii, Akihiko and Arandjelović, Relja and Sivic, Josef and Okutomi, Masatoshi and Pajdla, Tomas},
  booktitle={2015 IEEE Conference on Computer Vision and Pattern Recognition (CVPR)}, 
  title={24/7 place recognition by view synthesis}, 
  year={2015},
  volume={},
  number={},
  pages={1808-1817},
  doi={10.1109/CVPR.2015.7298790}}

@inproceedings{park2022simsac,
  title={Dual Task Learning by Leveraging Both Dense Correspondence and Mis-Correspondence for Robust Change Detection With Imperfect Matches},
  author    = {Jin-Man Park and
               Ue-Hwan Kim and
               Seon-Hoon Lee and
               Jong-Hwan Kim},
  year = {2022},
  booktitle = {2022 {IEEE} Conference on Computer Vision and Pattern Recognition, {CVPR} 2022}
}

@misc{bandara2022transformerbasedsiamesenetworkchange,
      title={A Transformer-Based Siamese Network for Change Detection}, 
      author={Wele Gedara Chaminda Bandara and Vishal M. Patel},
      year={2022},
      eprint={2201.01293},
      archivePrefix={arXiv},
      primaryClass={cs.CV},
      url={https://arxiv.org/abs/2201.01293}, 
}

@article{zhang2024bifa,
  title={Bifa: Remote sensing image change detection with bitemporal feature alignment},
  author={Zhang, Haotian and Chen, Hao and Zhou, Chenyao and Chen, Keyan and Liu, Chenyang and Zou, Zhengxia and Shi, Zhenwei},
  journal={IEEE Transactions on Geoscience and Remote Sensing},
  volume={62},
  pages={1--17},
  year={2024},
  publisher={IEEE}
}

@article{bandara2022ddpm,
  title={Ddpm-cd: Remote sensing change detection using denoising diffusion probabilistic models},
  author={Bandara, Wele Gedara Chaminda and Nair, Nithin Gopalakrishnan and Patel, Vishal M},
  journal={arXiv preprint arXiv:2206.11892},
  volume={3},
  year={2022},
  publisher={June}
}

@misc{teed2020raftrecurrentallpairsfield,
      title={RAFT: Recurrent All-Pairs Field Transforms for Optical Flow}, 
      author={Zachary Teed and Jia Deng},
      year={2020},
      eprint={2003.12039},
      archivePrefix={arXiv},
      primaryClass={cs.CV},
      url={https://arxiv.org/abs/2003.12039}, 
}

@misc{xu2022gmflowlearningopticalflow,
      title={GMFlow: Learning Optical Flow via Global Matching}, 
      author={Haofei Xu and Jing Zhang and Jianfei Cai and Hamid Rezatofighi and Dacheng Tao},
      year={2022},
      eprint={2111.13680},
      archivePrefix={arXiv},
      primaryClass={cs.CV},
      url={https://arxiv.org/abs/2111.13680}, 
}

@misc{huang2022flowformertransformerarchitectureoptical,
      title={FlowFormer: A Transformer Architecture for Optical Flow}, 
      author={Zhaoyang Huang and Xiaoyu Shi and Chao Zhang and Qiang Wang and Ka Chun Cheung and Hongwei Qin and Jifeng Dai and Hongsheng Li},
      year={2022},
      eprint={2203.16194},
      archivePrefix={arXiv},
      primaryClass={cs.CV},
      url={https://arxiv.org/abs/2203.16194}, 
}

@misc{sun2021loftrdetectorfreelocalfeature,
      title={LoFTR: Detector-Free Local Feature Matching with Transformers}, 
      author={Jiaming Sun and Zehong Shen and Yuang Wang and Hujun Bao and Xiaowei Zhou},
      year={2021},
      eprint={2104.00680},
      archivePrefix={arXiv},
      primaryClass={cs.CV},
      url={https://arxiv.org/abs/2104.00680}, 
}

@misc{edstedt2023romarobustdensefeature,
      title={RoMa: Robust Dense Feature Matching}, 
      author={Johan Edstedt and Qiyu Sun and Georg Bökman and Mårten Wadenbäck and Michael Felsberg},
      year={2023},
      eprint={2305.15404},
      archivePrefix={arXiv},
      primaryClass={cs.CV},
      url={https://arxiv.org/abs/2305.15404}, 
}

@Article{Chen2020,
AUTHOR = {Chen, Hao and Shi, Zhenwei},
TITLE = {A Spatial-Temporal Attention-Based Method and a New Dataset for Remote Sensing Image Change Detection},
JOURNAL = {Remote Sensing},
VOLUME = {12},
YEAR = {2020},
NUMBER = {10},
ARTICLE-NUMBER = {1662},
URL = {https://www.mdpi.com/2072-4292/12/10/1662},
ISSN = {2072-4292},
DOI = {10.3390/rs12101662}
}

@article{zhou2023unified,
  title={A unified deep learning network for remote sensing image registration and change detection},
  author={Zhou, Rufan and Quan, Dou and Wang, Shuang and Lv, Chonghua and Cao, Xianwei and Chanussot, Jocelyn and Li, Yi and Jiao, Licheng},
  journal={IEEE Transactions on Geoscience and Remote Sensing},
  volume={62},
  pages={1--16},
  year={2023},
  publisher={IEEE}
}
}

\end{document}